\documentclass{article}
\usepackage{arxiv}

\usepackage[utf8]{inputenc} % allow utf-8 input
\usepackage[T1]{fontenc}    % use 8-bit T1 fonts
\usepackage{hyperref}       % hyperlinks
\usepackage{url}            % simple URL typesetting
\usepackage{booktabs}       % professional-quality tables
\usepackage{amsfonts}       % blackboard math symbols
\usepackage{nicefrac}       % compact symbols for 1/2, etc.
\usepackage{microtype}      % microtypography
\usepackage{lipsum}
\usepackage{pifont}%
\usepackage{graphicx}
\usepackage{multirow}%
\usepackage{tabularx}%
\usepackage[title]{appendix}
\usepackage{amsmath,amssymb,amsfonts}%
\usepackage[table, dvipsnames]{xcolor}

\usepackage{natbib}

\newcolumntype{Y}{>{\centering\arraybackslash}X}

\title{Unsupervised Emotion Pattern Discovery in Long-Form Speech via Acoustic Clustering and Language Model Interpretation}

\author{
  Francesco Manco$^{\dag}$ \\
  Sidea Group \\
  \texttt{francesco.manco@sideagroup.com}
  \And
  Ivan Rinaldi$^{\dag}$ \\
  Department of Computer Science\\
  University of Bari Aldo Moro\\
  \texttt{i.rinaldi4@phd.uniba.it}
  \And
  Roberto Anglani \\
  Bip AI Lab Srl\\
  \texttt{roberto.anglani@bip-group.com}
  \And
  Eufemia Lella \\
  Sidea Group \\
  \texttt{eufemia.lella@sideagroup.com}
  \And
  Giovanna Castellano \\
  Department of Computer Science\\
  University of Bari Aldo Moro\\
  \texttt{giovanna.castellano@uniba.it}
  \And
  Gennaro Vessio \\
  Department of Computer Science\\
  University of Bari Aldo Moro\\
  \texttt{gennaro.vessio@uniba.it}
}

\begin{document}

\maketitle
\begin{abstract}
Speech emotion recognition is a key task in affective computing and multimedia analysis. However, most existing systems rely on supervised learning with predefined emotion categories and labeled data, making them unsuitable for long-form, real-world conversations where emotional states evolve over time and manual annotation is costly or impractical. In this work, we propose an unsupervised framework for emotion analysis in extended spoken interactions. The method integrates speaker diarization, acoustic feature extraction via a fine-tuned Wav2Vec2.0 model, and clustering techniques to automatically uncover latent emotional patterns without the need for labeled datasets. A distinctive contribution of our approach is the use of Large Language Models to interpret the resulting clusters, generating context-aware, human-readable descriptions that go beyond discrete emotion tags. We evaluate the system on multi-speaker conversations across diverse topics, comparing three analysis strategies: audio-only, text-only, and a hybrid approach that combines acoustic and linguistic cues. Results show that the hybrid method yields the most coherent and interpretable emotional analyses, preserving contextual consistency and supporting nuanced affective interpretations. The proposed framework offers a scalable and domain-adaptable solution for emotion-rich speech analysis, opening new opportunities in customer service, media monitoring, and empathic human-computer interaction. The code is publicly available at: \url{https://github.com/justivanr/unsupervised_emotion_discovery}.
\end{abstract}


\section{Introduction}

Human speech is a complex and expressive communication medium, conveying not only linguistic content but also rich paralinguistic cues such as emotion, intention, and affective states. While traditional approaches to speech processing have primarily focused on lexical and syntactic analysis, recent years have witnessed a growing interest in the emotional dimension of speech, with applications ranging from customer service and healthcare to media analysis and human-computer interaction.

Most existing speech emotion recognition (SER) systems rely on supervised learning techniques trained on labeled datasets (e.g., \cite{busso2008iemocap, cao2014crema, livingstone2018ryerson}). These approaches have proven effective in controlled scenarios involving short utterances and predefined emotion categories (e.g., \textit{joy}, \textit{sadness}, \textit{anger}). However, they suffer from several critical limitations when applied to real-world conversations: (\textit{i}) emotional states are often ambiguous, overlapping, or dynamic over time; (\textit{ii}) collecting and labeling large emotional speech corpora is costly and domain-specific; (\textit{iii}) discrete emotion labels offer limited interpretability, especially in long-form, multi-speaker interactions.

These limitations hinder the scalability, adaptability, and transparency of current emotion analysis systems, making them unsuitable for many practical scenarios. In particular, there is a lack of methodologies capable of handling spontaneous speech in its full temporal and contextual complexity and of providing interpretable outputs that go beyond simple categorical classification.

To address these challenges, we propose an \textit{unsupervised} and \textit{explainable} framework for emotion analysis in long-form speech. The proposed system begins with the segmentation and diarization of raw multi-speaker audio, allowing for the decomposition of conversations into coherent speaker-specific segments. These segments are then processed using a self-supervised speech model, specifically a fine-tuned Wav2Vec2.0 architecture \cite{baevski2020wav2vec}, to extract rich emotional embeddings from the acoustic signal. Without relying on predefined labels, clustering techniques are applied to these embeddings in order to uncover latent emotional structures and recurring affective patterns. Finally, Large Language Models (LLMs), including multimodal variants, are employed to interpret and describe the resulting clusters in natural language, thus enabling a semantically rich and human-understandable emotional summary of the conversation.

The framework is evaluated using real-world case studies that involve long, spontaneous conversations in Italian. We compare three cluster analysis approaches: acoustic-only, text-only, and a hybrid strategy that integrates both modalities. Quantitative results, supported by BERTScore~\cite{zhang2019bertscore} and CLAP~\cite{wu2023large} metrics, as well as qualitative evaluations, indicate that the hybrid method provides the most coherent and contextually relevant emotional insights.

The contributions of this work are as follows:
\begin{itemize}
    \item We introduce a fully unsupervised, domain-adaptable pipeline for emotion analysis in long-form speech, minimizing the need for manual labeling and enabling application in diverse scenarios.
    \item We propose to use a new dataset constructed by combining two existing datasets and augmenting them with newly generated samples using data augmentation techniques.
    \item We demonstrate the value of combining acoustic and textual information for interpretable emotion clustering, showing that the hybrid method significantly outperforms unimodal approaches.
    \item We evaluate the use of LLMs to explain acoustic emotion clusters in natural language, bridging the gap between low-level signal processing and high-level semantic understanding.
\end{itemize}
By integrating recent advances in self-supervised learning, clustering, and generative language models, our method moves beyond traditional emotion recognition to support a more nuanced, interpretable, and scalable form of affective speech analysis.

The remainder of the paper is organized as follows: Section~\ref{sec:related} reviews related work; Section~\ref{sec:materials} describes the datasets used and the data augmentation strategies adopted; Section~\ref{sec:method} presents the proposed framework; Section~\ref{sec:experiments} details the experimental setup and results; Section~\ref{sec:conclusion} concludes the paper and outlines directions for future research.

\section{Related Work}\label{sec:related}

Speech emotion recognition has traditionally relied on supervised learning techniques, which require large, manually annotated datasets with predefined emotional categories. For example, Bhattacharya et al.~\cite{bhattacharya2022emotion} tackled multilingual emotion recognition by augmenting labeled data and converting it into visual representations, which are then fed into a Convolutional Neural Network for classification. Kilimci et al.~\cite{kilimci2025evaluating} bypassed feature extraction entirely, using raw audio as input across six benchmark datasets---including EMO-DB~\cite{burkhardt2005database}, RAVDESS, and TESS~\cite{SP2/E8H2MF_2020}. Meanwhile, Heydari and Shahidi~\cite{heydari2025lightweight} proposed a lightweight architecture tailored for real-time emotion recognition on low-resource devices, leveraging both spectral and prosodic features.

Although these methods perform well on benchmark datasets, they primarily focus on high-resource languages such as English and German. 

Moreover, these systems are constrained by several key limitations, including the subjectivity of emotion annotations and the limited contextual depth provided by discrete emotional labels.

Recent research has explored multimodal approaches that combine acoustic and textual features to improve emotional understanding. For example, Jia and SungChu~\cite{jia2020deep} proposed a bimodal sentiment analysis system for real-world call center data, demonstrating the value of integrating speech and transcript information. In~\cite{garcia2021sentiment}, the authors addressed the challenge of variable-length audio inputs using Fully Convolutional Networks, which allow for more flexible and efficient processing without requiring prior segmentation.

The advent of self-supervised learning has significantly advanced SER, enabling the extraction of high-quality representations from raw speech without the need for manual annotation. Transformer-based models such as Wav2Vec2.0~\cite{baevski2020wav2vec}, HuBERT~\cite{hsu2021hubert}, and WavLM~\cite{chen2022wavlm} have demonstrated strong performance across diverse speech tasks. These models benefit from pretraining on large unlabeled corpora and exhibit robustness to noise, speaker variability, and domain shifts~\cite{jafarzadeh2024speaker,diatlova2024adapting}. Notably, Diatlova et al.~\cite{diatlova2024adapting} showed that incorporating speaker-specific metadata, such as gender, can further enhance performance. At the same time, the contribution of semantic features alone remains limited in comparison to acoustic cues.

Despite these advancements, most supervised SER methods still face three persistent challenges: poor generalization to unseen domains, dependency on labeled data, and limited interpretability. This underscores the need for alternative strategies that move beyond rigid classification schemes and support more flexible, explainable, and context-sensitive emotional analysis.

To this end, recent studies have begun integrating LLMs and multimodal LLMs (MLLMs) into the SER pipeline. SpeechCueLLM~\cite{wu2024beyond} has presented an innovative approach in which low-level acoustic features—such as tone, pitch, and intensity—are first translated into textual representations called \textit{impressions}. These impressions allow the LLM to reason about emotional states using its language understanding capabilities, without requiring direct audio input. Experiments on two different datasets show that this method enhances recognition accuracy, particularly in high-fidelity audio scenarios.

A complementary line of research is presented in OmniVox~\cite{murzaku2025omnivox}, which investigated the zero-shot performance of advanced MLLMs—such as Gemini 2.0 and GPT-4o—on SER tasks. The study demonstrates that, when equipped with well-designed prompts and enriched contextual information, these models can rival or surpass traditional supervised systems. Crucially, the authors emphasize the role of prompt engineering, highlighting how acoustic descriptors and conversational history contribute to more accurate and fine emotional inferences.

\begin{table}[!b]
    \centering
    \normalsize
    \caption{A comparative overview of state-of-the-art models and approaches in relation to our research objectives. Our method supports unsupervised learning, offers interpretable results, and enables targeted analysis, allowing users to focus on particular topics of interest.}
    \begin{tabular}{lccc}
        \toprule
        \textbf{Approach} & \textbf{Unsupervised} & \textbf{Interpretability} & \textbf{Targeted Analysis} \\
        \midrule  
        Jia \& SungChu~\cite{jia2020deep} & \ding{55} & \ding{55} & \ding{55} \\
        \midrule
        García-Ordás et al.~\cite{garcia2021sentiment} & \ding{55} & \ding{55} & \ding{55} \\
        \midrule
        Pepino et al.~\cite{pepino2021emotion} & \ding{55} & \ding{55} & \ding{55}  \\
        \midrule
        Jafarzadeh et al.~\cite{jafarzadeh2024speaker} & \ding{55} & \ding{55} & \ding{55} \\
        \midrule
        Diatlova et al.~\cite{diatlova2024adapting} & \ding{55} & \ding{55} & \ding{55} \\
        \midrule
        SpeechCueLLM~\cite{wu2024beyond} & \ding{51} & \ding{55} & \ding{55} \\
        \midrule
        OmniVox~\cite{murzaku2025omnivox} & \ding{51} & \ding{55} & \ding{55} \\
        \midrule
        \textit{Our Proposal} & \ding{51} & \ding{51} & \ding{51} \\
        \bottomrule
    \end{tabular}
    \label{tab:comparison_sota}
\end{table}

Together, these works mark a paradigm shift in SER—from purely feature-based emotion classification to semantically enriched, context-aware emotional understanding—made possible by the reasoning capabilities of foundation models. Table~\ref{tab:comparison_sota} summarizes the main characteristics of these approaches in relation to the goals of this study.

Nevertheless, critical limitations remain. Most of these systems still rely, directly or indirectly, on annotated data—either for model training, fine-tuning, or prompt evaluation—which poses significant scalability challenges in real-world settings. Furthermore, emotional outputs are often constrained to discrete categories, which limits interpretability and fails to capture the nuanced, evolving nature of affective dynamics in natural speech. Finally, existing frameworks typically operate on isolated speech segments, lacking mechanisms for analyzing emotions concerning specific topics, themes, or phases within long, multi-speaker conversations.

These open issues motivate the development of unsupervised, interpretable, and context-sensitive frameworks for analyzing long-form emotional speech—an objective that our proposed method aims to address.

\section{Materials}\label{sec:materials}

Most speech emotion recognition research to date has focused on English-language corpora, limiting the generalizability of models to other linguistic and cultural contexts. However, how emotions are verbally expressed is deeply influenced by the phonetic structure, prosodic patterns, and cultural conventions of each language. For instance, intonation, rhythm, and the use of specific affective expressions vary significantly across languages and can profoundly impact emotion perception.

In this study, we focus on the Italian language, which remains underrepresented in the SER literature despite its importance in key application domains such as healthcare, customer service, and media analysis. Developing and fine-tuning emotion recognition models for a specific language is essential to capture the nuanced ways in which emotions are expressed verbally and to ensure that SER technologies are effective and reliable in real-world contexts.

To this end, we selected two complementary datasets: \textit{EMOVO}~\cite{costantini2014emovo} and \textit{Emozionalmente}~\cite{catania2025emozionalmente}. These datasets were chosen based on their linguistic alignment, emotional coverage, and diversity in speaker profiles, making them suitable for robust, domain-adaptable SER.

\textit{}\textit{EMOVO} is the first emotional speech dataset explicitly designed for Italian. It features acted speech recorded by six professional actors (three male and three female), each performing 14 phonetically rich sentences across seven commonly used emotional categories: \textit{disgust}, \textit{fear}, \textit{anger}, \textit{joy}, \textit{surprise}, \textit{sadness}, and \textit{neutral}. This results in 98 utterances per actor and a total of 588 high-quality audio files. Despite its limited size compared to English-language corpora, \textit{EMOVO} remains a valuable resource for controlled SER tasks due to its balanced emotional distribution, textual consistency, and controlled acoustic conditions.

In contrast, \textit{Emozionalmente} is a large-scale dataset of emotional speech collected through a crowdsourcing approach. It includes 6,902 audio samples from 431 non-professional speakers who were asked to simulate the same seven basic emotions. This dataset introduces speaker diversity and greater acoustic variability, offering a more natural and spontaneous expression of emotional states—complementing the studio-recorded precision of \textit{EMOVO}.

To maximize both phonetic consistency and expressive variability, we combined these two datasets into a unified corpus, which we refer to as \textit{Emovonalmente}, consisting of 7,490 annotated utterances. This merged dataset enables the model to learn from both highly controlled and highly diverse speech conditions, improving its generalization capabilities in real-world scenarios. 

\begin{figure}[!b]
    \centering
    \includegraphics[width=.9\textwidth]{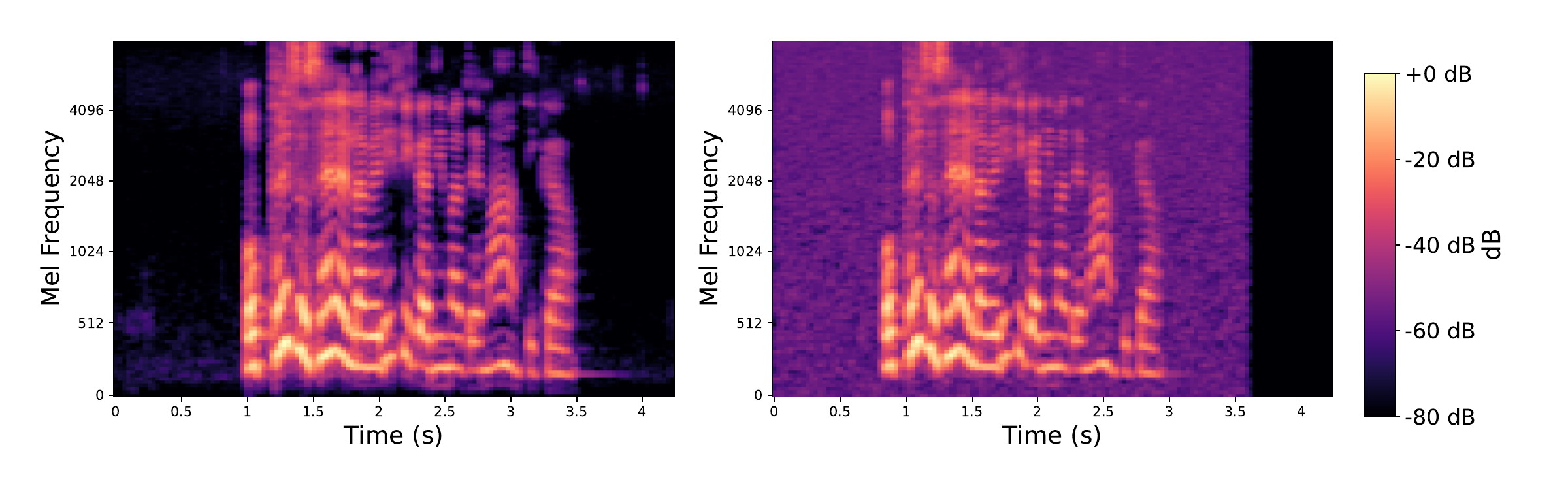} 
    \caption{Example of an audio segment's original mel-spectrogram (left) compared to the version after applying data augmentation techniques (right).}
    \label{fig:spect_augmentation_example}
\end{figure}

To further enhance the model's robustness, we applied data augmentation to the training data. The augmentations included the addition of Gaussian noise, time stretching, pitch shifting, time shifting, gain variation, and polarity inversion. These transformations simulate typical acoustic variations encountered in spontaneous speech, such as background noise, changes in speaking rate, and volume fluctuation. Figure~\ref{fig:spect_augmentation_example} provides a visual comparison of these transformations in the spectrogram domain. The goal of this augmentation strategy is to expose the model to a wide range of signal perturbations during training, thereby increasing its resilience to noise and improving its ability to perform reliably in uncontrolled acoustic environments.

\section{Methods}\label{sec:method}

\begin{figure}[!b]
  \centering
  \includegraphics[width=1.\textwidth]{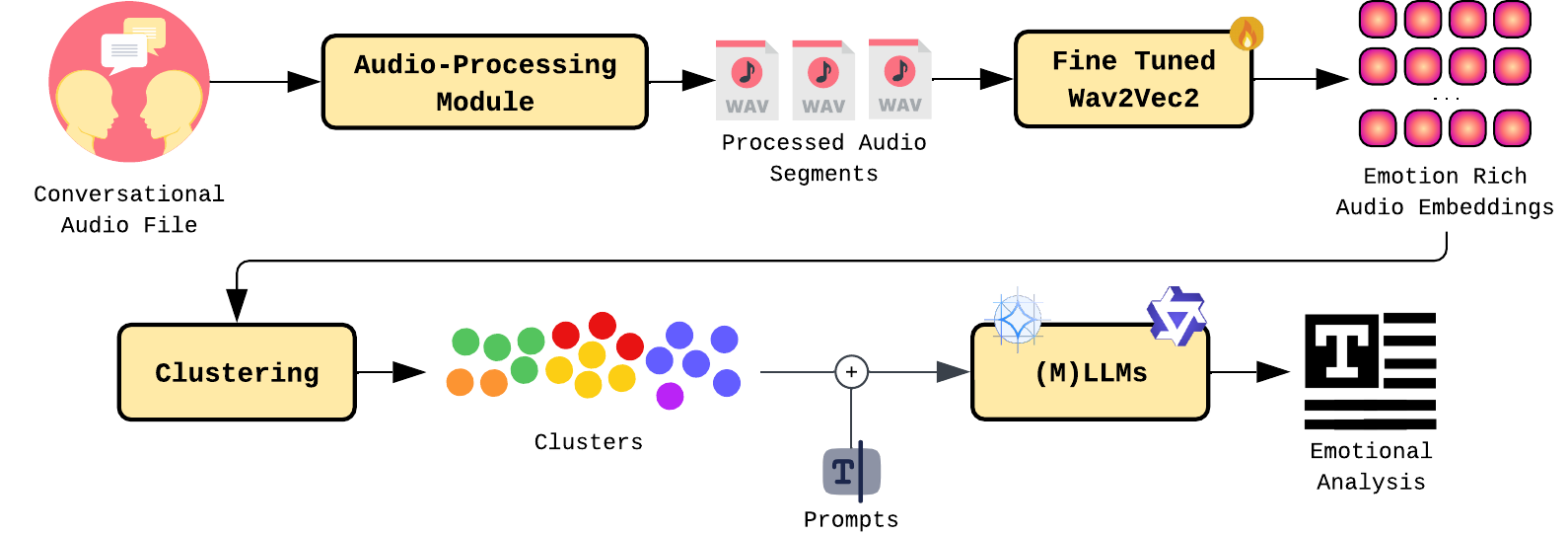}
  \caption{Overview of the proposed framework architecture. A long-form conversational audio input is first processed by the pre-processing module, which performs segmentation and speaker diarization to produce isolated audio segments. These segments are then fed into a Wav2Vec2.0 model fine-tuned for SER in Italian, from which emotional embeddings are extracted. The embeddings are clustered using an unsupervised algorithm to identify similar emotional expressions. Finally, the most representative segments from each cluster are passed to an LLM (or MLLM), which generates an interpretable emotional description summarizing the dominant emotions and other relevant information contained in the input audio.}
  \label{fig:hermes_architecture}
\end{figure}

The proposed framework for long-form, unsupervised speech emotion analysis is composed of four modular stages, each dedicated to a specific task: (\textit{i}) audio pre-processing and segmentation; (\textit{ii}) emotional feature extraction; (\textit{iii}) unsupervised clustering; and (\textit{iv}) semantic interpretation of emotional patterns. The overall architecture is illustrated in Fig.~\ref{fig:hermes_architecture}.

\subsection{Audio Pre-processing and Segmentation}

\begin{figure}[!b]
  \centering
  \includegraphics[width=1.\textwidth]{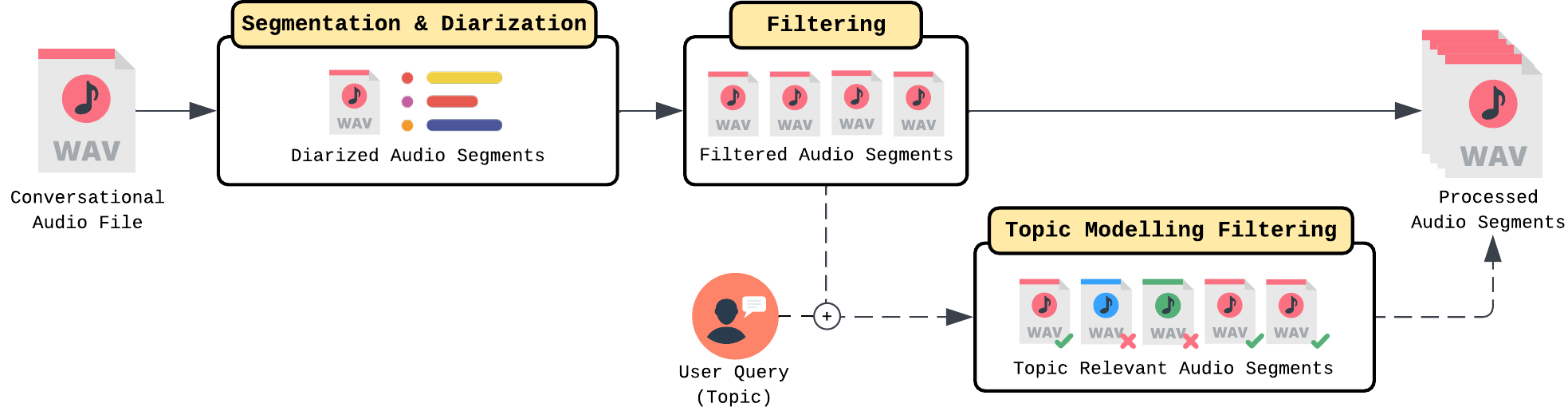}
    \caption{Audio pre-processing module of the proposed framework. The module supports two distinct processing modes. In the \textit{default mode} (indicated by solid arrows), the full conversation is processed end-to-end: all diarized segments are forwarded to the Wav2Vec2.0 model for emotional feature extraction. In the \textit{topic-focused mode} (indicated by dotted arrows), the analysis is narrowed based on a user-defined query. Only segments whose content is semantically related to the query are selected and passed to the subsequent stages of the framework.}
  \label{fig:hermes_proc_module}
\end{figure}

The first step involves preparing raw conversational audio for analysis. The pre-processing module, depicted in Fig.~\ref{fig:hermes_proc_module}, addresses practical challenges such as long audio durations, multiple speakers, and silent intervals. It supports two operational modes: a \textit{default mode}, which processes all available segments, and a \textit{topic-focused mode}, which filters content according to a user-specified topic of interest.

To handle multi-speaker interactions and improve downstream analysis, the input audio is segmented and diarized using the WhisperX model~\cite{bain2023whisperx}, selected for its robust voice activity detection, accurate word-level transcription, and speaker attribution. WhisperX transcribes and temporally aligns the audio, then applies speaker diarization to assign unique labels (e.g., ``Speaker A'', ``Speaker B'') to continuous speech segments from the same individual. An example output of this process is shown in Table~\ref{tab:diarization_example}.

\begin{table}[!t]
    \centering
    \caption{Example of diarized audio segments obtained after segmentation. Each entry includes start and end timestamps, speaker ID, transcription, and duration. These segments are used for emotional feature extraction and clustering.}
    \normalsize
    \begin{tabular}{ccclc}
        \toprule
        \textbf{Start (s)} & \textbf{End (s)} & \textbf{Speaker} & \textbf{Translation} & \textbf{Duration (s)} \\
        \midrule
        0.54 & 2.12 & Speaker A & 
        \begin{tabular}[t]{@{}l@{}}Ciao a tutti, come state?\\\textit{(Hello everyone, how are you?)}\end{tabular} & 1.58 \\
        2.31 & 3.89 & Speaker B & 
        \begin{tabular}[t]{@{}l@{}}Tutto bene, grazie.\\\textit{(All good, thanks.)}\end{tabular} & 1.58 \\
        4.05 & 7.15 & Speaker A & 
        \begin{tabular}[t]{@{}l@{}}Sono contento di vedervi qui.\\\textit{(Glad to see you here.)}\end{tabular} & 3.10 \\
        7.50 & 9.90 & Speaker B & 
        \begin{tabular}[t]{@{}l@{}}Anche noi, è un piacere.\\\textit{(We too, it is a pleasure.)}\end{tabular} & 2.40 \\
        10.15 & 13.50 & Speaker A & 
        \begin{tabular}[t]{@{}l@{}}Abbiamo molto di cui parlare oggi.\\\textit{(We have a lot to talk about today.)}\end{tabular} & 3.35 \\
        13.700 & 15.300 & Speaker B & 
        \begin{tabular}[t]{@{}l@{}}Certo, iniziamo subito.\\\textit{(Sure, let's start right away.)}\end{tabular} & 1.60 \\
        \bottomrule
    \end{tabular}
    \label{tab:diarization_example}
\end{table}

\begin{table}[!t]
    \centering
    \caption{Prompt template used to generate interpretable topic labels via LLM. We replace placeholders \texttt{[DOCUMENTS]} and \texttt{[KEYWORDS]} at runtime with relevant content extracted during topic modeling.}
    \normalsize
    \begin{tabular}{l>{\raggedright\arraybackslash}p{10.5cm}}
        \toprule
        User Prompt & 
        Ho un argomento che contiene i seguenti documenti: [DOCUMENTS] \newline
        L'argomento è descritto dalle seguenti parole chiave: [KEYWORDS] \newline
        In base alle informazioni sopra, estrai un'etichetta tematica breve \newline
        ma molto descrittiva, di massimo due parole. \newline
        Assicurati che sia nel seguente formato: \texttt{topic:\ etichetta tematica} \newline
        \textit{(I have a topic that contains the following documents: [DOCUMENTS] \newline
        The topic is described by the following keywords: [KEYWORDS] \newline
        Based on the information above, extract a short \newline
        but very descriptive subject label of maximum two words. \newline
        Make sure it is in the following format: \texttt{topic:\ thematic label})}\\
        \midrule
        Model Output & \texttt{topic:\ Sfida Intellettuale} \textit{(\texttt{topic:\ Intellectual Challenge})} \\
        \bottomrule
    \end{tabular}
    \label{tab:BERT_TOPIC_prompt}
\end{table}

\begin{table}[!t]
    \centering
    \caption{Examples of topics identified by the topic modeling module. Each topic is described by its ID, number of associated segments, generated label, and an excerpt from a representative transcription.}
    \normalsize
    \begin{tabular}{ccll}
        \toprule
        \textbf{ID} & \textbf{Count} & \textbf{Topic Label} & \textbf{Representative Excerpt} \\
        \midrule
        1 & 171 & 
        \begin{tabular}[t]{@{}l@{}}Dinamiche Familiari\\\textit{(Family Dynamics)}\end{tabular} & 
        \begin{tabular}[t]{@{}l@{}}mi ricordo benissimo tutto di mio padre e tut\ldots \\
        \textit{(I remember perfectly everything about my father and all\ldots)} \end{tabular} \\
        \cmidrule{1-4}
        2 & 152 & 
        \begin{tabular}[t]{@{}l@{}}Riconoscimento Personale\\\textit{(Personal Recognition)}\end{tabular} & 
        \begin{tabular}[t]{@{}l@{}}soddisfazione di dire anch'io ti ho amata ti\ldots\\\textit{(satisfaction of saying I loved you too\ldots)}\end{tabular} \\
        \cmidrule{1-4}
        3 & 153 & 
        \begin{tabular}[t]{@{}l@{}}Libri Proibiti\\\textit{(Forbidden Books)}\end{tabular} & 
        \begin{tabular}[t]{@{}l@{}}c'era un librone dei libri proibiti dove tutt\ldots\\
        \textit{(there was a big book of forbidden books where all\ldots)}\end{tabular} \\
        \cmidrule{1-4}
        4 & 144 & 
        \begin{tabular}[t]{@{}l@{}}Famiglia e Ricordi\\\textit{(Family and Memories)}\end{tabular} & 
        \begin{tabular}[t]{@{}l@{}}Mio padre era severo, che comprò mio nonno, c\ldots\\
        \textit{(My father was strict, who bought my grandfather, c\ldots)}\end{tabular} \\
        \bottomrule
    \end{tabular}
    \label{tab:extracted_topics_example}
\end{table}

Segments are then filtered based on their duration, retaining only those between 1.5 and 10 seconds in length. Short adjacent segments from the same speaker are merged to form emotionally meaningful units, while overly long ones are split at natural midpoints to maintain linguistic and emotional coherence. Specifically, to handle audio segments longer than 10 seconds, we employ a midpoint-based segmentation strategy: the midpoint of the segment is computed (e.g., at 6 seconds for a 12-second segment), and the word temporally closest to it is selected as the split location. The segment is then divided into two parts at this word boundary. To ensure meaningful segmentation, we verify that each resulting subsegment contains at least two words; otherwise, the original segment remains unsplit. This approach prevents the creation of trivial or degenerate segments that may arise from long pauses or sparse speech, even in temporally extended utterances.

In the topic-focused mode, an additional sub-module selects only the segments relevant to a specific theme. This is achieved using BERTopic~\cite{grootendorst2022bertopic}, which operates on transcriptions embedded via multilingual SentenceTransformer models~\cite{reimers2019sentence}, dimensionally reduced using a custom UMAP~\cite{mcinnes2018umap} configuration, and clustered via K-Means~\cite{macqueen1967multivariate}. Topic refinement is performed using class-based TF-IDF (c-TF-IDF), and final topic labels are generated using Gemma3~\cite{team2025gemma} for enhanced interpretability. Table~\ref{tab:BERT_TOPIC_prompt} outlines the prompting strategy, and Table~\ref{tab:extracted_topics_example} reports sample extracted topics.

\subsection{Feature Extraction via Wav2Vec2.0}

Filtered audio segments are passed to a feature extraction module based on a fine-tuned Wav2Vec2.0 model. The model was trained on Italian emotional speech data to learn rich, high-level embeddings that encode affective information without requiring predefined emotion labels. The resulting embeddings are temporally pooled to obtain fixed-size vectors, summarizing the emotional content of each segment. These embeddings retain both acoustic and latent emotional features, enabling downstream clustering.

\subsection{Unsupervised Emotion Clustering}

The extracted embeddings are input to a clustering module that identifies groups of speech segments with similar emotional characteristics. As the framework operates in a fully unsupervised setting, clustering enables the discovery of latent emotional patterns not tied to predefined categories. Clusters serve as the basis for later interpretation and explanation via LLMs or MLLMs.

\subsection{Interpretation via LLMs and MLLMs}
To provide human-readable descriptions of each emotional cluster, we explore three distinct interpretation strategies:

\begin{itemize}
    \item \textit{Textual approach}: Based solely on transcriptions of representative audio segments.
    \item \textit{Audio-based approach}: Relies on (M)LLMs to directly interpret raw audio segments.
    \item \textit{Hybrid approach}: Enriches transcriptions with qualitative descriptions of acoustic properties.
\end{itemize}

In the hybrid approach, we compute five acoustic features commonly associated with emotional expression: average volume, volume variation, pitch, pitch variation, and speaking rate. These are then converted into textual impressions, following the SpeechCueLLM methodology~\cite{wu2024beyond}, to aid the LLM in capturing paralinguistic cues.

To guide the generation of emotional summaries, we designed specific system prompts for each interpretation modality—text-only, hybrid (text + acoustic description), and audio-only. For the text-only approach, the prompt instructs the LLM to analyze a cluster of transcripts and return a structured summary comprising: the primary emotion with intensity, three keywords capturing the emotional essence, a concise emotion label (e.g., \textit{joy}, \textit{anger}), a brief explanation justifying the emotion based on the textual content, and the main topic of the cluster. For the hybrid approach, the structure remains the same. However, the LLM also receives additional acoustic descriptions for each segment (e.g., tone, pace, volume), enabling it to combine textual and prosodic cues in the emotional interpretation. For the audio-only approach, the prompt is adapted for use with an MLLM, which receives raw audio inputs. The model must extract shared emotional characteristics from the audio clips, again following a five-field output: primary emotion with intensity, keywords, emotion label, explanation (based on prosody), and main topic. In all three cases, we enforce a consistent and constrained output format to enhance comparability and facilitate evaluation. The specific prompts used in our experiments are available in our public repository, which will be made accessible upon acceptance.

The model is fed with the $K$ most representative segments per cluster, selected according to a custom \textit{audio representativeness} score:
\begin{equation}
audrep_i = w_d \cdot (1 - d_{i,\text{norm}}) + w_t \cdot t_{i,\text{norm}}.
\label{eq:representativeness}
\end{equation}
Here, $d_{i,\text{norm}}$ and $t_{i,\text{norm}}$ are the min-max normalized distance from the cluster centroid and the duration of the $i$-th segment, respectively. The weights $w_d$ and $w_t$ (with $w_d + w_t = 1$) control the trade-off between closeness to the centroid and richness of content. This score prioritizes emotionally representative segments that are sufficiently long to carry semantic and affective information.

For each cluster, the LLM (or MLLM) generates a structured summary that includes:
\begin{itemize}
    \item A \textit{dominant emotional category} or \textit{state}.
    \item \textit{Keywords} associated with the segment content.
    \item A \textit{brief explanatory rationale} for the detected emotion.
    \item An optional \textit{label} for the main topic being discussed.
\end{itemize}

\section{Experiments}\label{sec:experiments}

This section presents the experimental setup used to evaluate the proposed framework. We describe the models and hyperparameter configurations, outline the performance metrics, and report the results of our experiments.

\subsection{Experimental Setting}\label{sec:exp_setting}

All fine-tuning and inference operations were performed on an AWS virtual machine equipped with an NVIDIA L4 GPU with 24 GB of VRAM.

We fine-tuned the Wav2Vec2.0 model \cite{grosman2021xlsr53-large-italian}, pre-trained on the Italian subset of the \textit{Common Voice} dataset \cite{ardila2019common}. The adopted architecture consists of: (\textit{i}) the base Wav2Vec2.0 feature extractor; (\textit{ii}) a linear projection layer; and (\textit{iii}) a classification head mapping the extracted features to emotion classes. During training, the convolutional feature extractor was kept frozen, while the projection and classification layers were fine-tuned.

Fine-tuning was conducted on the \textit{Emovonalmente} dataset (see Sec.~\ref{sec:materials}), with data augmentation techniques applied to improve generalization. The resulting dataset comprises 10{,}784 Italian audio samples, split into 7{,}764 for training, 863 for validation, and 2{,}157 for testing (based on an 80/20 stratified train-test split, with 10\% of the training set used for validation). We employed the AdamW optimizer with a learning rate of $2 \cdot 10^{-5}$, weight decay of $0.001$, and a cosine learning rate scheduler with a warmup ratio of 0.1. Training was run for 10 epochs using a batch size of 8 and 2 gradient accumulation steps. Early stopping with a patience of 3 epochs was applied to prevent overfitting. Further details are provided in Section~\ref{sec:wav2vec2ft_ablation}.

Acoustic embeddings were clustered using K-Means with $K = 7$ and random initialization. For visualization and analysis purposes, UMAP was applied for dimensionality reduction, using a minimum distance of $0.0$, $10$ components, and $5$ neighbors. A grid search was performed to optimize the number of clusters, initialization method, and UMAP parameters. Motivations behind the use of K-Means over alternative clustering algorithms are discussed in Sec.~\ref{sec:clustering_ablation}.

To identify representative segments within each cluster, we computed the \textit{audrep} score (Eq.~\ref{eq:representativeness}), which combines normalized distance from the cluster centroid with segment duration. We empirically set the weights to $w_d = 0.6$ and $w_t = 0.4$, balancing prototypicality and content richness. A higher $w_d$ favors shorter but more emotionally salient segments, while a higher $w_t$ favors longer, possibly noisier, segments.

For emotional description generation, we employed the following generative models:
\begin{itemize}
    \item Gemma3-12B:it-q4\_K\_M, available via Ollama~\cite{ollama}, used in both the \textit{text-only} and \textit{hybrid} settings.
    \item Qwen2.5-Omni-7B, used for the \textit{audio-only} setting.
\end{itemize}

In the \textit{text-only} setting, Gemma3 received the transcripts of the top 10 representative segments per cluster, along with a prompt requesting identification of the dominant emotion based on linguistic cues.

In the \textit{hybrid} setting, the same segments were enriched with prosodic features (e.g., pitch, volume, articulation rate), according to the SpeechCueLLM framework, enabling Gemma3 to incorporate both textual and acoustic cues.

In the \textit{audio-only} setting, Qwen2.5-Omni was provided with 3 audio files per cluster, along with a prompt requesting emotional description. The rationale for using only three audio files is discussed in Sec.~\ref{sec:audioonly_ablation}.

\subsection{Performance Metrics}\label{sec:perf_metrics}

We adopt different evaluation metrics for each component of the framework.

For the fine-tuned Wav2Vec2.0 model, we report standard classification metrics: macro-averaged precision, recall, and F1-score. While informative, these metrics serve primarily as internal indicators and are not the main focus of the analysis.

To assess the quality of the clustering, we use the \textit{silhouette score}, which evaluates intra-cluster cohesion and inter-cluster separation:
\begin{equation}
\text{S}(i) = \frac{b(i) - a(i)}{\max(a(i), b(i))},
\label{eq:silhouette_score}
\end{equation}
where $a(i)$ is the average distance between sample $i$ and the other samples in the same cluster, and $b(i)$ is the average distance between $i$ and the samples in the nearest neighboring cluster.

To measure the alignment between generated emotional descriptions and their respective inputs (text or audio), we employ the following metrics.

BLEU (Bilingual Evaluation Understudy)~\cite{papineni2002bleu} measures $n$-gram precision between generated and reference texts:
\begin{equation}
\text{BLEU} = \text{BP} \cdot \exp\left(\sum_{n=1}^{N} w_n \log p_n\right),
\label{eq:bleu_score}
\end{equation}
where $\text{BP}$ is the brevity penalty, $w_n$ are the weights for each $n$-gram level, and $p_n$ is the precision for $n$-grams.

ROUGE (Recall-Oriented Understudy for Gisting Evaluation)~\cite{lin2004rouge} emphasizes recall. We report ROUGE-1 and ROUGE-L:
\begin{equation}
\text{ROUGE-1} = \frac{\text{Number of overlapping unigrams}}{\text{Number of unigrams in the reference}},
\label{eq:rouge1}
\end{equation}
\begin{equation}
\text{ROUGE-L F1} = \frac{(1 + \beta^2) \cdot \text{Recall} \cdot \text{Precision}}{\text{Recall} + \beta^2 \cdot \text{Precision}},
\label{eq:rougeL_f1}
\end{equation}
where ROUGE-L is based on the longest common subsequence between the candidate and the reference.

BERTScore~\cite{zhang2019bertscore} evaluates semantic similarity using contextual embeddings from BERT~\cite{devlin2019bert}. It computes Recall ($R_{\text{BERT}}$), Precision ($P_{\text{BERT}}$), and F1 as:
\begin{align}
R_{\text{BERT}} &= \frac{1}{|x|} \sum_{x_i \in x} \max_{\hat{x}_j \in \hat{x}} \mathbf{x}_i^\top \hat{\mathbf{x}}_j,
\label{eq:bertscore_recall} \\
P_{\text{BERT}} &= \frac{1}{|\hat{x}|} \sum_{\hat{x}_j \in \hat{x}} \max_{x_i \in x} \mathbf{x}_i^\top \hat{\mathbf{x}}_j,
\label{eq:bertscore_precision} \\
F_{\text{BERT}} &= 2 \cdot \frac{P_{\text{BERT}} \cdot R_{\text{BERT}}}{P_{\text{BERT}} + R_{\text{BERT}}}.
\label{eq:bertscore_f1}
\end{align}

CLAP Score~\cite{wu2023large} quantifies the semantic alignment between audio and text in a joint embedding space:
\begin{equation}
\text{CLAP Score} = \frac{\vec{e}_{\text{audio}} \cdot \vec{e}_{\text{text}}}{\|\vec{e}_{\text{audio}}\| \cdot \|\vec{e}_{\text{text}}\|},
\label{eq:clap_score}
\end{equation}
where $\vec{e}_{\text{audio}}$ and $\vec{e}_{\text{text}}$ denote the respective audio and text embeddings.

\subsection{Results}\label{sec:results}

\subsubsection{Wav2Vec2.0 Fine-Tuning}

\begin{table}[!t]
    \centering
    \caption{Performance of the fine-tuned Wav2Vec2.0 model on the test set across seven emotion categories. Metrics include precision, recall, and F1-score.}
    \normalsize
    \begin{tabular}{lccc}
        \toprule
        \textbf{Emotion} & \textbf{Precision} & \textbf{Recall} & \textbf{F1-score} \\
        \midrule
        Fear       & 0.792 & 0.748 & 0.769 \\
        Anger      & 0.864 & 0.893 & 0.878 \\
        Disgust    & 0.873 & 0.804 & 0.837 \\
        Sadness    & 0.769 & 0.822 & 0.795 \\
        Joy        & 0.803 & 0.776 & 0.791 \\
        Surprise   & 0.795 & 0.869 & 0.830 \\
        Neutrality & 0.919 & 0.897 & 0.908 \\
        \midrule
        \textbf{Macro Avg} & 0.831 & 0.830 & 0.830 \\
        \bottomrule
    \end{tabular}
    \label{tab:wav2vec_ft_res}
\end{table}

Table~\ref{tab:wav2vec_ft_res} reports the performance of the fine-tuned Wav2Vec2.0 model on the test set, with metrics computed per emotion category. The best-performing classes are \textit{anger}, \textit{disgust}, and \textit{neutrality}, while \textit{fear} and \textit{joy} exhibit relatively lower scores. Overall, the model demonstrates strong generalization across all seven emotion labels, achieving a macro F1-score of 0.830.

Further analysis reveals that the model performs particularly well on \textit{neutrality} and \textit{anger}, likely due to their distinctive acoustic profiles. Conversely, \textit{joy} and \textit{surprise} are frequently confused, possibly due to overlapping prosodic patterns. Similarly, instances of \textit{sadness} are often misclassified as \textit{fear} or \textit{disgust}, reflecting similar vocal characteristics. Examples of model predictions are presented in Table~\ref{tab:classification_examples}, including both correctly classified and misclassified samples.

\begin{table}[!t]
    \centering
    \caption{Examples of emotion classification using the fine-tuned Wav2Vec2.0 model. \textcolor{green}{\textbf{Green}} labels indicate correct predictions, while \textcolor{red}{\textbf{red}} labels indicate misclassifications. GT: ground truth. Pred: predicted label. Audio IDs specify whether the sample originated from the Emozionalmente (Emoz.) or EMOVO dataset.}
    \normalsize
    \begin{tabular}{llcc}
        \toprule
        \textbf{Audio ID} & \textbf{Transcription} & \textbf{GT} & \textbf{Pred} \\
        \midrule
        1613493660293 (Emoz.) & 
        \begin{tabular}[t]{@{}l@{}}Gli operai si alzano presto\\\textit{(The workers get up early)}\end{tabular} 
        & sadness & \textcolor{green}{\textbf{sadness}} \\

        1615737916558 (Emoz.) & 
        \begin{tabular}[t]{@{}l@{}}L’ho incontrato oggi dopo due anni\\\textit{(I met him today after two years)}\end{tabular} 
        & fear & \textcolor{green}{\textbf{fear}} \\

        m3/neu-m3-b2 (EMOVO) & 
        \begin{tabular}[t]{@{}l@{}}I vigili sono muniti di pistola\\\textit{(The police are equipped with guns)}\end{tabular} 
        & neutrality & \textcolor{green}{\textbf{neutrality}} \\

        1620376608487 (Emoz.) &
        \begin{tabular}[t]{@{}l@{}}\`E impegnato in una riunione\\\textit{(He's busy in a meeting)}\end{tabular} 
        & neutrality & \textcolor{red}{\textbf{anger}} \\

        m3/pau-m3-d1 (EMOVO) & 
        \begin{tabular}[t]{@{}l@{}}Sabato sera cosa farà?\\\textit{(What will you do on Saturday night?)}\end{tabular} 
        & fear & \textcolor{red}{\textbf{surprise}} \\
        \bottomrule
    \end{tabular}
    \label{tab:classification_examples}
\end{table}

\subsubsection{Clustering}

To assess the effectiveness of the learned representations for unsupervised emotion analysis, we conducted a twofold qualitative evaluation of the final-layer embeddings produced by the fine-tuned Wav2Vec2.0 model.

\begin{figure}[!t]
    \centering
    \includegraphics[width=0.8\textwidth]{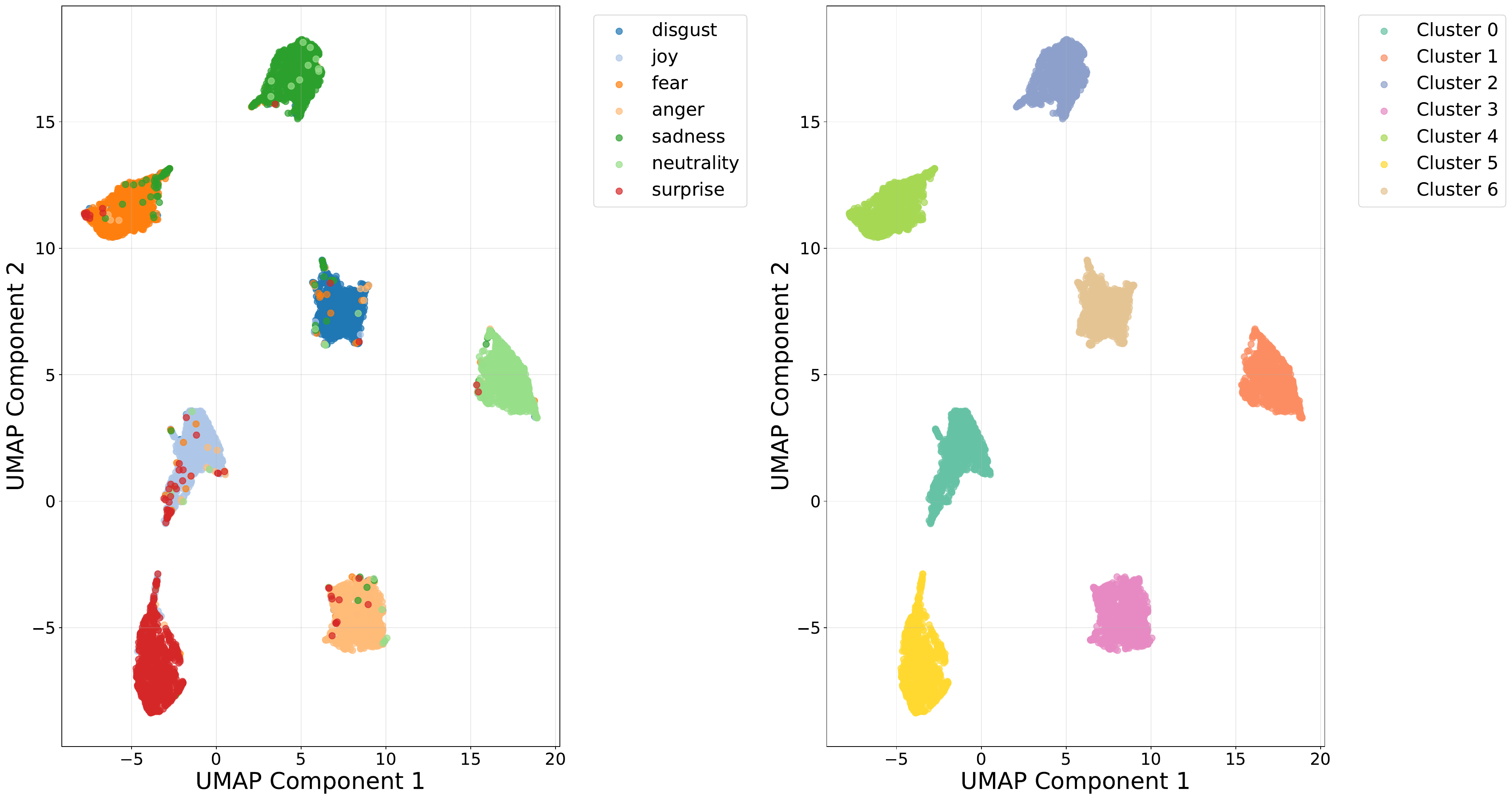}
    \caption{UMAP visualization of the final-layer audio embeddings. Left: embeddings colored by ground-truth emotion labels, illustrating the separability of emotional categories in the learned latent space. Right: clusters produced by K-Means ($k = 7$), highlighting the model's internal representation and unsupervised grouping of emotions.}
    \label{fig:umap_embeddings}
\end{figure}

First, we analyzed the latent space derived from the \textit{Emovonalmente} dataset. Figure~\ref{fig:umap_embeddings} shows a UMAP projection of the extracted embeddings, comparing ground-truth emotion labels with the clustering output produced by K-Means ($k = 7$). The observed separation among emotional categories demonstrates the model’s capacity to serve not only as a classifier but also as a meaningful feature extractor for downstream clustering tasks.

\begin{figure}[!t]
    \centering
    \includegraphics[width=0.8\linewidth]{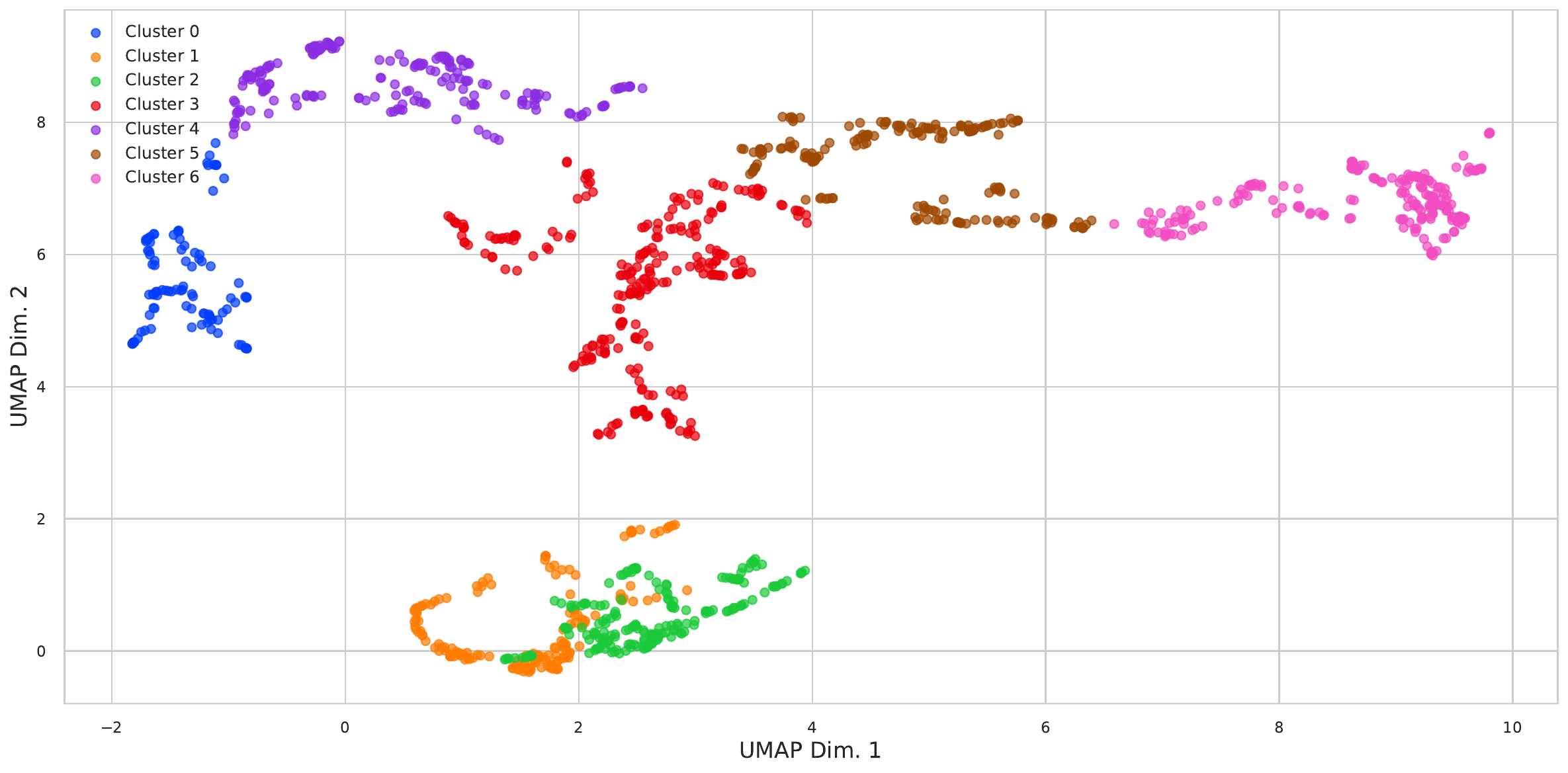}
    \caption{UMAP projection of K-Means clustering results on podcast embeddings. Each point represents an audio segment, colored by cluster assignment. The plot illustrates the internal structure and separation between clusters in the reduced latent space.}
    \label{fig:k_means_cluster_GL}
\end{figure}

To further evaluate the generalization of the framework in a real-world setting, we extracted audio embeddings from a publicly available Italian podcast episode on YouTube, with a duration of approximately 1 hour and 30 minutes. These embeddings were clustered using K-Means and projected via UMAP into two dimensions. The resulting structure, depicted in Fig.~\ref{fig:k_means_cluster_GL}, shows that the model also organizes unlabelled, in-the-wild audio segments into semantically coherent groups. The visualization reveals a clear separation among most clusters, suggesting that the learned embeddings effectively capture semantically or acoustically relevant variations within the podcast content. This confirms the suitability of combining Wav2Vec2.0 with UMAP and K-Means for unsupervised structuring of long-form audio data. Nonetheless, partial overlap between some clusters (e.g., Cluster 1 and Cluster 2) may reflect content continuity or similar prosodic patterns across adjacent speech segments. Despite these local ambiguities, the overall clustering outcome supports the model’s ability to identify meaningful distinctions in real-world audio streams.

\subsubsection{Emotional Analysis}
Lastly, we leveraged the generated clusters to guide the generation of emotional descriptions. We conducted a quantitative analysis across all clusters extracted from the podcast, we also provide qualitative insights to illustrate the framework’s output, including both the input data and the corresponding emotional descriptions.

\begin{table}[!b]
  \centering
  \caption{Average quantitative results for the three evaluated approaches: text-only, hybrid, and audio-only. The metrics assess semantic alignment either between the generated descriptions and the original transcripts (text-only and hybrid) or between the audio signals and the generated descriptions (audio-only). The best results for each metric are highlighted in bold. Abbreviations: BESc = BERTScore (F1, Precision, Recall); RG = ROUGE (RG1 = ROUGE-1, RGL = ROUGE-L); CLAPSc = CLAP Score.}
  \normalsize
  \begin{tabularx}{1\linewidth}{lYYYYYYY}
    \toprule
    \textbf{Approach} & \textbf{BLEU$\uparrow$} & \textbf{RG1$\uparrow$} & \textbf{RGL$\uparrow$} & \textbf{BEScF1$\uparrow$} & \textbf{BEScP$\uparrow$} & \textbf{BEScR$\uparrow$} & \textbf{CLAPSc$\uparrow$}\\
    \midrule
    Text-Only & \textbf{0.0074} & \textbf{0.1247} & \textbf{0.0801} & 0.5846 & 0.5970 & 0.5731 & - \\
    Hybrid & 0.0003 & 0.0723 & 0.0620 & \textbf{0.5930} & \textbf{0.6046} & \textbf{0.5817} & - \\
    Audio-Only & - & - & - & - & - & - & \textbf{0.4197} \\
    \bottomrule
    \end{tabularx}%
  \label{tab:avg_quant_metrics}
\end{table}

We first conducted a quantitative evaluation using metrics suited to each data type, text in the text-only and hybrid approaches, and audio in the audio-only approach.
Table~\ref{tab:avg_quant_metrics} summarizes the average performance across the three approaches. For the text-only and hybrid settings, coherence was evaluated using BLEU, ROUGE, and BERTScore, comparing the generated descriptions against the original transcripts. The text-only method obtained higher BLEU and ROUGE scores, indicating better surface-level similarity. However, the hybrid approach achieved superior semantic alignment, as reflected by its higher BERTScore values (e.g., an F1 score of $0.5930$ compared to $0.5846$ for the text-only method). This suggests that the integration of acoustic features with textual content enhances the model's understanding of the underlying emotional message, resulting in more informative and accurate descriptions.

For the audio-only approach, we employed CLAP Score to measure the alignment between the generated text and the corresponding audio input. Although this method produced lower average scores compared to the text-based settings, this outcome is expected, given the lack of direct linguistic input. Nonetheless, the audio-only model was effective in capturing salient paralinguistic and affective cues—such as tone, mood, and atmosphere—which may be crucial in scenarios where text is unavailable or unreliable. These results reinforce our findings: the hybrid approach yields the most comprehensive understanding by combining semantic and acoustic cues.

We then carried out a qualitative analysis to evaluate each method's ability to interpret emotional content. This involved manually reviewing the representative segments and comparing them to the descriptions generated by each model. We present detailed results for two representative clusters—Cluster 4 and Cluster 5—to provide qualitative insights into the framework’s output.

\begin{table}[!b]
  \centering
  \caption{Comparison of emotional descriptions for Cluster~4 generated by the three modalities: text-only, hybrid, and audio-only. Each entry reports the main emotion, relevant keywords, the assigned emotional label, a brief explanatory note, and the inferred topic.}
  \small
  \begin{tabular}{p{0.2\textwidth}p{0.75\textwidth}}
    \toprule
    \textbf{Modality} & \textbf{Generated description} \\
    \midrule
    \textbf{Text-only (Gemma-3)} & 1. Cluster emotion: Sense of challenge and professional pride, with moderate pressure.
    
    2. Keywords: competition, pride, responsibility.
    
    3. Emotional label: Ambition 
    
    4. Explanation: The fragments reflect a complex work environment, consisting of responsibility, competition (chef vs. restaurateur, Michelin stars), ambition (international expansion, training at Savoy), and a sense of achievement (“success”). The need to “take the plunge” and manage pressure indicates a constant challenge.
    
    5. Main topic: Professional growth and management of complex goals in the sector. \\
    \midrule
    \textbf{Hybrid (Gemma-3)} & 1. Main emotion: Strong sense of commitment and anxiety about professional success.
    
    2. Keywords: Commitment, success, anxiety, management.
    
    3. Emotional label: Ambition
    
    4. Explanation: The fragments show a strong focus on business management, achieving success, and pressure to maintain high standards, with frequent references to responsibility and commitment. The speed of speech and variations in volume suggest intense concentration and a certain degree of anxiety related to these goals. 
    
    5. Main topic: The speakers reflect on the management, growth, and challenges of a business. \\
    \midrule
    \textbf{Audio-only (Qwen2.5-Omni)} & 1. Main Emotion: Stress - High.
    
    2. Keywords: responsibility, collision, overview.
    
    3. Emotional Label: Pressure.
    
    4. Explanation: The speaker feels pressured due to the responsibilities and the need to manage multiple aspects in a restaurant setting.
    
    5. Main Topic: Restaurant management. \\
    \bottomrule
  \end{tabular}
  \label{tab:desc_cluster_4}
\end{table}

\begin{table}[!t]
  \centering
  \caption{Emotional descriptions generated for Cluster~5 across different modalities, including identified emotions, key expressions, explanatory notes, and inferred topics.}
  \small
  \begin{tabular}{p{0.2\textwidth}p{0.75\textwidth}}
    \toprule
    \textbf{Modality} & \textbf{Generated description} \\
    \midrule
    \textbf{Text-only (Gemma-3)} & 1. Cluster emotion: Nostalgic reflection and mild satisfaction, moderately intense.
    
    2. Keywords: Nostalgia, pride, journey.
    
    3. Emotional label: Memory.
    
    4. Explanation: The fragments exhibit a reflective tone, recounting past experiences (London, work, people) with a sense of pride and slight regret, typical of nostalgia. The lack of salary demands, professional training, and the recounting of a journey suggest an appreciation of the past. 
    
    5. Main topic: Positive review of past work experiences and relationships. \\
    \midrule
    \textbf{Hybrid (Gemma-3)} & 1. Cluster emotion: Amazement and reflection mixed with energy. 
    
    2. Keywords: surprise, energy, reflection.
    
    3. Emotional label: Wonder.
    
    4. Explanation: The fragments often express unexpected reactions and personal memories, often related to work or travel experiences. The variation in tone, volume, and speed of speech, combined with the audio descriptions, suggests a dynamic emotional response, ranging from surprise to enthusiasm, complemented by moments of reflection.
    
    5. Main topic: Memories of past experiences that inspire wonder and personal growth. \\
    \midrule
    \textbf{Audio-only (Qwen2.5-Omni)} & 1. Main Emotion: Surprise - High.
    
    2. Keywords: surprised, unexpected, proficiency.
    
    3. Emotional Label: Surprised.
    
    4. Explanation: The speaker is surprised by the high proficiency of someone in a certain field, which is unexpected.
    
    5. Main Topic: Learning and proficiency. \\
    \bottomrule
  \end{tabular}

  \label{tab:desc_cluster_5}
\end{table}

Specifically, we show both the input data and the corresponding emotional descriptions for these two clusters. As shown in Table~\ref{tab:desc_cluster_4}, the text-only description for Cluster~4 labeled the theme as ``Ambizione'' (``Ambition''), inferred from keywords such as ``responsabilità'' and ``successo'' (``responsibility'' and ``success''). However, this approach failed to capture the emotional tension conveyed through vocal tone. The audio-only method instead labeled the emotion as ``Pressure'', based on prosodic features like rapid speech and a steady pitch contour. While this is a plausible interpretation, it lacks semantic grounding. In contrast, the hybrid approach synthesized both perspectives, preserving the ambition theme while enriching it with affective nuances, such as ``commitment and anxiety''—inferred from variations in volume and speech rate.

Cluster~5 illustrates even more striking differences. As reported in Table~\ref{tab:desc_cluster_5}, the text-only method identified the theme ``Ricordo'' (``Memory''), based on repeated references to personal names (e.g., Luca, Marco), suggesting a narrative of reminiscence. The audio-only method interpreted the emotion as ``Surprise'', likely triggered by intonational peaks. However, this may have been influenced by a few salient acoustic frames rather than the overall content. The hybrid method, instead, proposed the label ``Meraviglia'' (``Astonishment''): while not perfectly accurate, this interpretation proved more nuanced and semantically expressive than the previous ones. By combining acoustic and textual cues, the hybrid approach delivered a richer and more balanced emotional understanding. Ideally, one could envision a future strategy that merges the topic-level coherence of text-based models with the emotional depth provided by multimodal integration.

\section{Conclusion}\label{sec:conclusion}

In this paper, we proposed a framework for analyzing emotional dynamics in long, spontaneous conversations without relying on manually labeled data. Our approach integrates a fine-tuned Wav2Vec2.0 model—adapted for Italian emotional speech—with unsupervised clustering and Large Language Models to interpret emerging emotional patterns. The main finding is that a hybrid strategy, combining acoustic and textual features, produces more accurate, nuanced, and interpretable emotional descriptions than unimodal alternatives. We also introduced a new dataset, \textit{Emovonalmente}, obtained by merging the EMOVO and Emozionalmente corpora and applying data augmentation techniques to increase linguistic and emotional diversity. This resource enables fine-tuning of speech emotion recognition models specifically for Italian and offers a solid foundation for future research in the field. Experimental evaluation across multiple podcast episodes—featuring diverse speakers and topics—demonstrated the framework’s robustness and adaptability. While the experimental results focus on a single podcast episode for illustrative purposes, the framework was also tested on multiple other episodes and consistently demonstrated robustness and adaptability. 

Looking ahead, potential enhancements include refining the preprocessing pipeline through improved diarization and transcription, exploring few-shot prompting to boost emotional expressiveness, and conducting systematic studies on prompt engineering to increase output quality and control. Further research could also assess the performance of different LLMs and MLLMs for emotional analysis, aiming to identify the most effective configurations. Finally, incorporating additional evaluation metrics would allow for a more comprehensive assessment of the relevance, coherence, and affective depth of the generated descriptions with respect to the input signals.

\section*{Author Contributions}
All authors contributed to the conceptualization of the study and the development of the methodology. FM and IR implemented the code and conducted the experiments. FM, IR, and GV prepared the first draft of the manuscript. All remaining authors reviewed and provided critical feedback on the draft. RA, EL, GC, and GV supervised the overall research activity.

\section*{Funding}
Partial funding for this work was provided by Regione Puglia through the EVOCITY project, co-funded under the Programma Regionale Puglia FESR–FSE+ 2021–2027 programme (Project Code: CDP001523, CUP: B95H24000470007).

Ivan Rinaldi received financial support from a PhD fellowship awarded within the framework of the project ``Territorial Pacts of Higher Education for Enterprises'', part of Mission 4, Component 2, Investment 3.3, co-funded by AI Factory S.r.l.\ (CUP: F61B23000370006).

\bibliographystyle{plain}
\bibliography{references}

\begin{appendices}

\section{Ablation Studies}\label{sec:ablation}

\subsection{Wav2Vec2.0 Fine-Tuning} \label{sec:wav2vec2ft_ablation}

Table~\ref{tab:ft_ablation_res} reports the results of fine-tuning Wav2Vec2.0 on different dataset configurations. We first evaluated the original EMOVO and Emozionalmente datasets, then combined them into a unified dataset, \textit{Emovonalmente}, and applied various data augmentation techniques to assess their impact. Each dataset was split using an 80/20 train-test ratio, with a further 90/10 split on the training set to create a validation subset. Data augmentation was applied only to the final training portion.

For example, the \textit{Emovonalmente} dataset (7,490 instances) was split into 5,992 training and 1,498 test samples; then, the training set was further divided into 5,392 training and 600 validation samples. In the \textit{on-the-fly} setting, augmentation is dynamically performed while loading the audio data during training, keeping the number of instances constant while varying the input audio at each epoch. The best model was selected based on the F1-macro score.

\begin{table}[!t]
    \centering
    \caption{F1-macro scores and dataset sizes for various fine-tuning configurations. We compare the original (as-is), fully augmented (full aug.), and on-the-fly versions of EMOVO, Emozionalmente, and their combination (Emovonalmente). ``Full aug.'' indicates data augmentation applied to the training subset. ``On-the-fly'' refers to augmentation applied dynamically when loading the data.}
    \begin{tabular}{lrc}
        \toprule
        \textbf{Dataset} & \textbf{Instances} & \textbf{F1-macro} \\
        \midrule
        EMOVO as-is & 588 & 0.166 \\
        EMOVO full aug. & 846 & 0.669 \\
        Emozionalmente as-is & 6,902 & 0.825 \\
        Emozionalmente full aug. & 9,936 & 0.836 \\
        Emovonalmente as-is & 7,490 & 0.820 \\
        Emovonalmente full aug. & 10,784 & 0.850 \\
        Emovonalmente on-the-fly & 7,490 & 0.820 \\
        \bottomrule
    \end{tabular}
    \label{tab:ft_ablation_res}
\end{table}

\begin{table}[!t]
    \centering
    \caption{Hyperparameters tested during Wav2Vec2.0 fine-tuning.}
    \begin{tabular}{lll}
        \toprule
        \textbf{Category} & \textbf{Hyperparameter} & \textbf{Value} \\
        \midrule
        \multirow{3}{*}{\textbf{Optimization}} 
            & Optimizer & AdamW \\
            & Learning Rate & $2\mathrm{e}{-5}$ \\
            & Weight decay & 0.001 \\
        \midrule
        \multirow{2}{*}{\textbf{Scheduler}} 
            & Type & Cosine \\
            & Warmup ratio & 0.1 \\
        \midrule
        \multirow{5}{*}{\textbf{Training}} 
            & Epochs & [10, 20, 30] \\
            & Batch size & 8 \\
            & Grad.~accum.~steps & 2 \\
            & Early stopping patience & [3, 4, 5] \\
            & Best model metric & \texttt{f1\_macro} \\
        \bottomrule 
    \end{tabular}
    \label{tab:hyperparams_finetuning}
\end{table}

While the \textit{Emozionalmente full aug.} dataset, where "full aug." means the entire training data was augmented, performed well ($F1 = 0.836$), the \textit{Emovonalmente full aug.} configuration yielded the highest score ($F1 = 0.850$). This improvement is attributed to increased diversity and better emotional coverage obtained by combining datasets and applying augmentation to a richer training set. Notably, the on-the-fly method matched the baseline Emovonalmente score ($0.820$), suggesting that it is a viable option when disk or preprocessing resources are limited. However, it underperforms compared to offline augmentation.

Based on these findings, we selected the \textit{Emovonalmente full aug.} dataset for all subsequent experiments. Table~\ref{tab:hyperparams_finetuning} summarizes the tested hyperparameters used during Wav2Vec2.0 fine-tuning. The best results were achieved using: \textit{Emovonalmente full aug.} (10,784 audio samples), AdamW with learning rate $2\mathrm{e}{-5}$ and weight decay 0.001, cosine scheduler with 0.1 warmup ratio, batch size 8, 2 gradient accumulation steps, and early stopping with patience set to 3 (maximum 10 epochs).

\subsection{Clustering Algorithms Comparison}  \label{sec:clustering_ablation}

Table~\ref{tab:grid_all_algorithms} lists all hyperparameters tested via grid search for UMAP and the clustering algorithms (K-Means, Agglomerative, and Spectral). We tested 216 configurations for K-Means, 270 for Agglomerative, and 108 for Spectral clustering.

\begin{table}[!t]
    \centering
    \caption{Grid of hyperparameters for UMAP and clustering algorithms.}
    \begin{tabular}{ll}
        \toprule
        \textbf{Component} & \textbf{Hyperparameter} \\
        \midrule
        \multirow{3}{*}{\textbf{UMAP}} 
            & \texttt{n\_neighbors}: [5, 15, 30] \\
            & \texttt{min\_dist}: [0.0, 0.1] \\
            & \texttt{n\_components}: [2, 5, 10] \\
        \midrule
        \multirow{2}{*}{\textbf{K-Means}} 
            & \texttt{n\_clusters}: [5--10] \\
            & \texttt{init}: [k-means++, random] \\
        \midrule
        \multirow{2}{*}{\textbf{Agglomerative}} 
            & \texttt{n\_clusters}: [5--9] \\
            & \texttt{linkage}: [ward, complete, average] \\
        \midrule
        \multirow{3}{*}{\textbf{Spectral}} 
            & \texttt{n\_clusters}: [5--10] \\
            & \texttt{affinity}: nearest neighbors \\
            & \texttt{n\_neighbors}: 15 \\
        \bottomrule
    \end{tabular}
    \label{tab:grid_all_algorithms}
\end{table}

We evaluated clustering quality using the Silhouette Score (Eq.~\ref{eq:silhouette_score}). For each algorithm, we analyzed how the average score varied with the number of clusters ($K$), selecting the most stable configuration among the top-performing ones.

\subsection{Audio-Only Setting Limitations} \label{sec:audioonly_ablation}

In the \textit{audio-only} setting, we tested Qwen2.5-Omni with the same prompt structure as other modalities, using ten audio files in Italian. However, several limitations emerged:

\begin{itemize}
    \item The model tended to focus disproportionately on a few inputs while ignoring others. Reducing the number of files slightly improved coverage, but explicit prompts were still required.
    \item Outputs often contained spelling errors, repeated keywords, and semantically inconsistent phrases. Some responses were incomplete or incoherent. English prompts led to more stable results.
\end{itemize}

To mitigate these issues, we concatenated the ten files into three segments (each $<30$s) with 1–2s pauses between utterances. While this improved coherence slightly and complied with Qwen2.5-Omni’s input limits, the model still struggled to generate accurate, emotionally rich, and balanced descriptions. Further investigations are required to understand and resolve these limitations.

\end{appendices}

\end{document}